\documentclass[11pt]{article}

\usepackage[final]{acl}

\usepackage{times}
\usepackage{latexsym}
\usepackage[T1]{fontenc}
\usepackage[utf8]{inputenc}
\usepackage{microtype}
\usepackage{graphicx}
\usepackage{subcaption}
\usepackage{xcolor}
\usepackage{booktabs}
\usepackage{makecell}
\usepackage{xurl}
\usepackage{lipsum}

\title{Induction in Both Directions: A Mechanistic Analysis of In-Context Learning in Masked Diffusion Language Models}

\author{Andy Catruna\textsuperscript{1,2} \quad Emilian
  Radoi\textsuperscript{1,2} \\
      \textsuperscript{1}National University of Science and Technology
  POLITEHNICA Bucharest \\
      \textsuperscript{2}Academy of Romanian Scientists \\
      \texttt{\{andy\_eduard.catruna, emilian.radoi\}@upb.ro}}

\begin{document}
\maketitle

\begin{abstract}
While the internal mechanisms of autoregressive (AR) transformers have been studied extensively, much less is known about diffusion language models (DLMs), an emerging alternative that generates text by iterative denoising. In this work, we study how DLMs implement induction, a mechanism behind in-context learning in which the model finds a repeated context and copies the token that followed it. Our analysis compares attention-only AR models and absorbing-mask DLMs with matched architectures. We find that DLMs learn a bidirectional induction circuit, where previous-token and next-token heads write local context into the residual stream and later induction heads use it to find and copy the answer from the matching source position. The circuit is direction-symmetric, working whether the source appears in the past or in the future. When only left context is visible, matching what an AR model sees, the DLM does not outperform its AR counterpart in induction capabilities. However, we observe it has stronger induction when both sides of the masked token are visible, pointing to bidirectional context access rather than a stronger one-sided mechanism. Beyond induction, we provide causal evidence that DLMs compute the global fraction of masked tokens and use it as an implicit timestep, even though they are given no explicit timestep embedding.
\end{abstract}

\section{Introduction}
Large language models (LLMs) based on autoregressive (AR) transformers have become central tools for writing, coding and scientific work~\citep{noy2023experimental,lee2022coauthor,peng2023impact,boiko2023autonomous}, making their internal mechanisms an important object of study~\citep{bereska2024mechanistic}. However, AR transformers are not the only relevant class of language models. Diffusion language models (DLMs) are a growing alternative that generate text by repeatedly denoising sequences rather than predicting only the next token~\citep{austin2021structured,sahoo2024simple}. This gives them a different computational structure and the promise of parallel decoding, as shown by recent systems such as LLaDA~\citep{nie2026large}, DiffusionGemma~\citep{odonoghue2026diffusiongemma}, and Mercury~\citep{khanna2025mercury}.

\begin{figure}[t!]
\centering
\includegraphics[width=\linewidth]{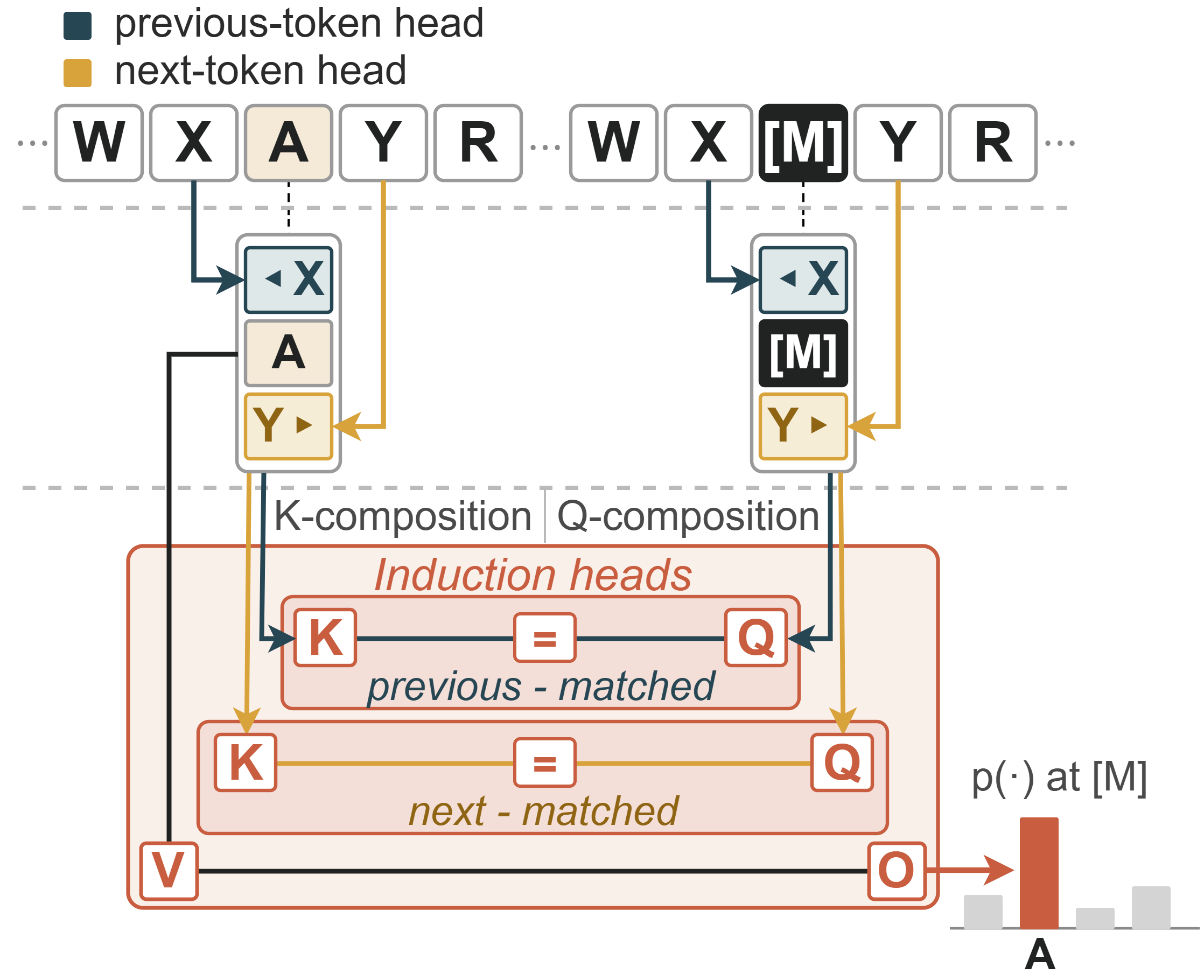}
\caption{The bidirectional induction circuit identified in DLMs. Neighbour heads write previous-token and next-token cues. Later induction heads use their QK weights to match those cues between the mask and the source answer, and their OV path raises the probability of the answer token at the mask. The same circuit applies when the source answer is in the future.}
\label{fig:circuit}
\end{figure}

AR language models have received extensive attention in mechanistic interpretability~\citep{bereska2024mechanistic,elhage2021mathematical,olsson2022context,wang2022interpretability,conmy2023towards}, but much less is known about the internal mechanisms of DLMs~\citep{wang2026dlm,dai2026revealing,kong2026mechanism}. As DLMs use a different training objective, bidirectional attention and an iterative denoising process, mechanisms found in AR models may not transfer directly~\citep{kong2026mechanism}.

In AR models, induction heads are a canonical circuit for in-context learning~\citep{olsson2022context}. They allow the model to use a repeated context to copy the token that followed the same context earlier, providing a basic mechanism for learning from examples inside the prompt. Understanding whether DLMs implement induction in the same way is a first step toward comparing their in-context learning mechanisms with those of AR models.

DLMs change the setting in which induction has to operate. They predict masked tokens from bidirectional context, but the true token at the prediction position is hidden behind the mask token \(\mathtt{[M]}\) (often written as \(\mathtt{[MASK]}\))~\citep{austin2021structured,sahoo2024simple}. Therefore, a DLM could use the same AR induction circuit, extend it symmetrically to copy from both past and future contexts, or use a different mechanism.

In this work, we study this problem in a controlled setting, focusing on absorbing-mask DLMs. We train matched attention-only AR transformers and DLMs, evaluate on the same repeated-token induction task, and analyze the DLM circuit with causal ablations and weight decompositions. This lets us compare behaviour and mechanism under the same architecture and data distribution.

Following standard practice in mechanistic interpretability~\citep{elhage2021mathematical, olsson2022context}, we use small models to enable circuit-level causal analysis, providing a foundation for understanding the mechanisms of large-scale DLMs.

Our analysis reveals that DLMs learn induction and implement it in a bidirectional form, as shown in Figure~\ref{fig:circuit}. Induction emerges abruptly, requires at least two layers, and works approximately equally well from past or future context.

This work makes the following contributions:
\begin{itemize}
    \item We identify and provide causal evidence for a two-pathway circuit in which previous-token and next-token heads write local context, and later induction heads use this information to copy from both past and future contexts.
    %\item We compare AR and DLM induction in a matched setting. When only left context is available, matching what an AR model sees, the DLM does not outperform its AR counterpart, and it has stronger induction only when both sides of the masked token are visible.
    \item We compare induction in AR models and DLMs under matched conditions and show that DLMs outperform AR models in terms of induction only when they can exploit bidirectional context, and not when restricted to the left-context setting used by AR models.
    %\item We show that DLMs outperform AR models only when they can exploit bidirectional context. Through a controlled comparison under matched contextual conditions, we find that DLMs exhibit no induction advantage when restricted to the same left context as AR models.
    \item We show that DLMs encode the global mask fraction as an implicit timestep without having an explicit timestep embedding. A linear probe can recover the global mask fraction from the residual stream at a single position after the first attention layer, and patching the mask-rate direction causally changes prediction entropy.
\end{itemize}

\section{Related Work}

\textbf{Mechanistic Interpretability of Transformer Language Models.} Autoregressive transformers have been the main focus of mechanistic interpretability in language models. Prior work introduced tools for analyzing transformer circuits, including residual-stream decomposition, attention-head circuits, and QK/OV factorizations~\citep{elhage2021mathematical}. This line of work identified induction heads as a central mechanism for in-context learning in AR models, where earlier tokens provide the features needed to copy from repeated contexts~\citep{olsson2022context}.

Mechanistic interpretability has also developed causal and representation-level methods for studying model behaviour~\citep{mueller2026quest}. Circuit analyses and path patching~\citep{wang2022interpretability,goldowsky2023localizing,conmy2023towards} test which components are necessary for a behaviour, while probing and sparse-feature methods~\citep{belinkov2022probing,huben2024sparse,templeton2026scaling} study what information is represented in the residual stream. Work on superposition further shows why individual neurons may not correspond cleanly to single features~\citep{elhage2022toy}. Our DLM analysis draws on several of these approaches, combining mean ablations, QK/OV decomposition, and linear probing.

\textbf{Diffusion Language Models.} DLMs produce text by gradually denoising corrupted sequences instead of generating tokens one by one from left to right. Early work developed discrete diffusion objectives, including absorbing-state and more general transition processes, and later work adapted these objectives to language modeling~\citep{austin2021structured,gulrajani2023likelihood,lou2023discrete,sahoo2024simple,shi2024simplified}. We use DLM as the broad term in this paper, and our experiments focus on absorbing-mask DLMs, which prior work often calls masked diffusion models (MDMs)~\citep{ou2025your,zheng2025masked}.

Recent work has made DLMs more competitive by scaling masked diffusion models, adapting AR models into diffusion models, and combining autoregressive and diffusion-style generation~\citep{nie2025scaling,gong2025scaling,arriola2025block}. Large DLMs further show that diffusion-based generation is becoming a practical alternative to standard AR decoding~\citep{nie2026large,ye2025dream,khanna2025mercury,odonoghue2026diffusiongemma}. These models are often motivated by faster or more parallel decoding, but their bidirectional denoising objective~\citep{artetxe2022role,patel2022bidirectional} also changes the internal computation for in-context learning~\citep{kim2025train}.

\textbf{Interpretability of Masked Diffusion Models.} Mechanistic work on masked diffusion models is still emerging. Recent studies analyze sparse features in DLMs, attention-floating behaviour during denoising, mechanism changes when AR models are post-trained into masked diffusion models, and theoretical links between masked diffusion and learned-order autoregression~\citep{wang2026dlm,dai2026revealing,kong2026mechanism,garg2025masked}. These papers show that DLMs can share some behaviour with AR models while also changing how information is routed. Our work instead identifies the DLM induction circuit and provides causal evidence for its two-pathway structure in a matched AR-DLM setting.

\section{Method}

\subsection{Matched Autoregressive and Diffusion Models}

We train matched attention-only AR and DLM transformers at depths $L\in\{1,2,3\}$. The AR models use causal attention and next-token prediction, whereas the DLMs use bidirectional attention and an absorbing-mask diffusion objective~\citep{austin2021structured,sahoo2024simple}. Apart from the objective and attention mask, the architecture and training recipe are the same for both types of models. For the main behavioural comparisons, we train three independent seeds for each family and depth, allowing us to compare both induction emergence and final induction performance across model families.

\subsection{Measuring Induction}
We evaluate induction capabilities on sequences constructed with random tokens where only in-context information can be utilized to predict the correct tokens~\citep{olsson2022context}. Each sequence has length 512, with the first 256 tokens sampled uniformly at random and repeated as the second half. Let \(\mathtt{W\,X\,A\,Y\,R}\) denote a five-token source window with answer token \(\mathtt{A}\), and let \(\mathtt{W\,X\,[M]\,Y\,R}\) denote the corresponding query window with \(\mathtt{A}\) replaced by the mask token \(\mathtt{[M]}\). In forward induction, the source is in the first copy and the query in the second, whereas in reverse induction their order is swapped. In total, we use 128 such sequences for evaluating each model. We mask a fixed, deterministic set of non-adjacent positions and score 3,328 masked tokens for the forward evaluation and 3,328 masked tokens for the reverse evaluation.

We use random tokens so that semantic and natural-language frequency effects cannot explain performance and the model must use the in-context information to identify the masked token. We score induction by how much repetition helps the model predict the correct token: for each masked position, we compare the model's log probability on the correct token in the repeated sequence to its log probability on the same token in a non-repeated control sequence with the same mask layout. The induction score is this difference in nats, averaged over masked positions.

AR models have no mask token \(\mathtt{[M]}\), so we evaluate them with next-token prediction on the same sequences: the model reads the sequence without any masks and predicts each scored position from the tokens to its left, using the same position-matched control. This makes the AR score directly comparable to the DLM forward score, while the reverse setting has no AR counterpart.

\subsection{Circuit Analysis}
In order to localize the circuits, we utilize two complementary measurements. First, we compute structural attention fingerprints, such as how much a head attends from a masked position $i$ to $i-1$, to $i+1$, or to the source answer position in the other copy, which has no mask tokens. Second, we perform mean ablations: for each head, we replace its output with its per-position mean over non-repeated control prompts and recompute the induction score~\citep{wang2022interpretability}. This approach helps us identify which heads attend to relevant information as well as which are actually causally important for induction.

For weight-level analysis, we use DLMs without LayerNorm, obtained by converting models trained with LayerNorm, so that the residual stream can be decomposed into additive components. For QK analysis, we decompose an induction head's attention logit from \(\mathtt{[M]}\) to the source answer into query-component by key-component terms. Following \citet{elhage2021mathematical}, a head uses Q-composition when an earlier head's output contributes to its query, and K-composition when it contributes to its key. For OV analysis, we pass source-token embeddings through an induction head's value and output matrices and then through the unembedding, giving an effective token-to-token logit map.

\subsection{Training and Implementation Details}
All models use $d_{\mathrm{model}}=512$, 16 attention heads with $d_{\mathrm{head}}=32$, and sequence length 512. The DLM objective uses $T=100$ timesteps and no explicit timestep embedding. Models are trained on FineWeb-Edu sequences~\citep{penedo2024fineweb} for 250k steps with batch size 128, learning rate $10^{-3}$, 10k warmup steps, and cosine decay to 10\% of the peak learning rate. We intentionally train in the overtraining regime commonly used in mechanistic interpretability work~\citep{olsson2022context,nanda2023progress}. After training, AR models have lower validation perplexity at every depth (e.g., $64.5$ vs.\ $117.5$ at L2), although the DLM value is an ELBO-based upper bound on the true perplexity~\citep{sahoo2024simple}, so the gap may be smaller than these numbers suggest. However, we observe that in some settings, DLM induction is stronger.

Both model families use the GPT-2 BPE tokenizer~\citep{radford2019language}, and for DLMs we add one extra mask token \(\mathtt{[M]}\). We train all models with AdamW~\citep{loshchilov2017decoupled} with weight decay 0.1 and gradient clipping at 1.0. For DLMs, each batch samples a timestep $t\in\{1,\ldots,T\}$ uniformly and independently masks each position with cosine probability $1-\cos((t/T)\pi/2)$. We average the loss only over masked positions.

Behavioural results are measured on AR and DLM models trained with LayerNorm. For mechanistic analysis, we use folded no-LayerNorm DLMs, which we obtain as follows~\citep{nanda2022transformerlens,heimersheim2024you}: we fully train the DLMs with LayerNorm, then replace each LayerNorm's per-token standard deviation with a constant calibrated on training data. This makes every LayerNorm a linear map, which we fold into the adjacent weight matrices. Finally, we briefly fine-tune the resulting no-LayerNorm models until they recover the performance of the LayerNorm models.

This transformation approximately preserves model quality: at L2, the folded no-LayerNorm DLMs reach nearly the same validation perplexity as the LayerNorm models ($117.2$ vs.\ $117.5$), and the folded L2 and L3 models show the same induction circuit. Their induction scores stay strong but are not identical: forward scores are $3.67 \pm 0.15$ nats at L2 and $4.64 \pm 0.09$ at L3 (Table~\ref{tab:replication}), versus $4.13 \pm 0.17$ and $4.47 \pm 0.31$ for the LayerNorm models.

\section{Results}
We first compare induction behaviour in AR and DLM models, then localize the heads that implement the DLM circuit and test their causal role. Afterward, we analyze how their QK and OV weights find and copy the source answer token, and finally show that DLMs compute and use the global mask rate as an implicit timestep.

\begin{figure}[hbt]
\centering
\begin{subfigure}{\linewidth}
\centering
\includegraphics[width=\linewidth]{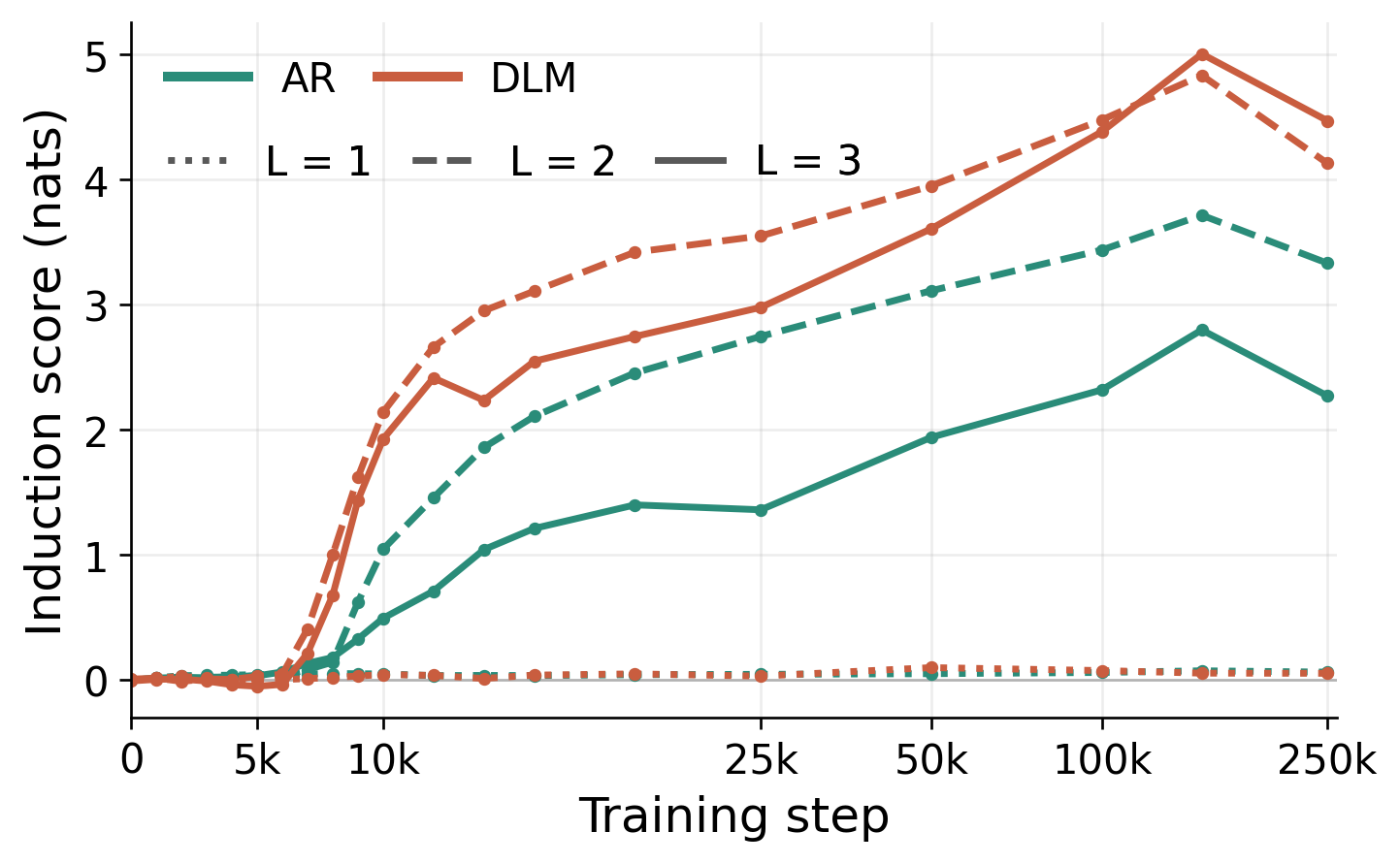}
\caption{Induction emerges abruptly and peaks around 150k steps.}
\label{fig:phase-change}
\end{subfigure}
\par\smallskip
\begin{subfigure}{\linewidth}
\centering
\includegraphics[width=\linewidth]{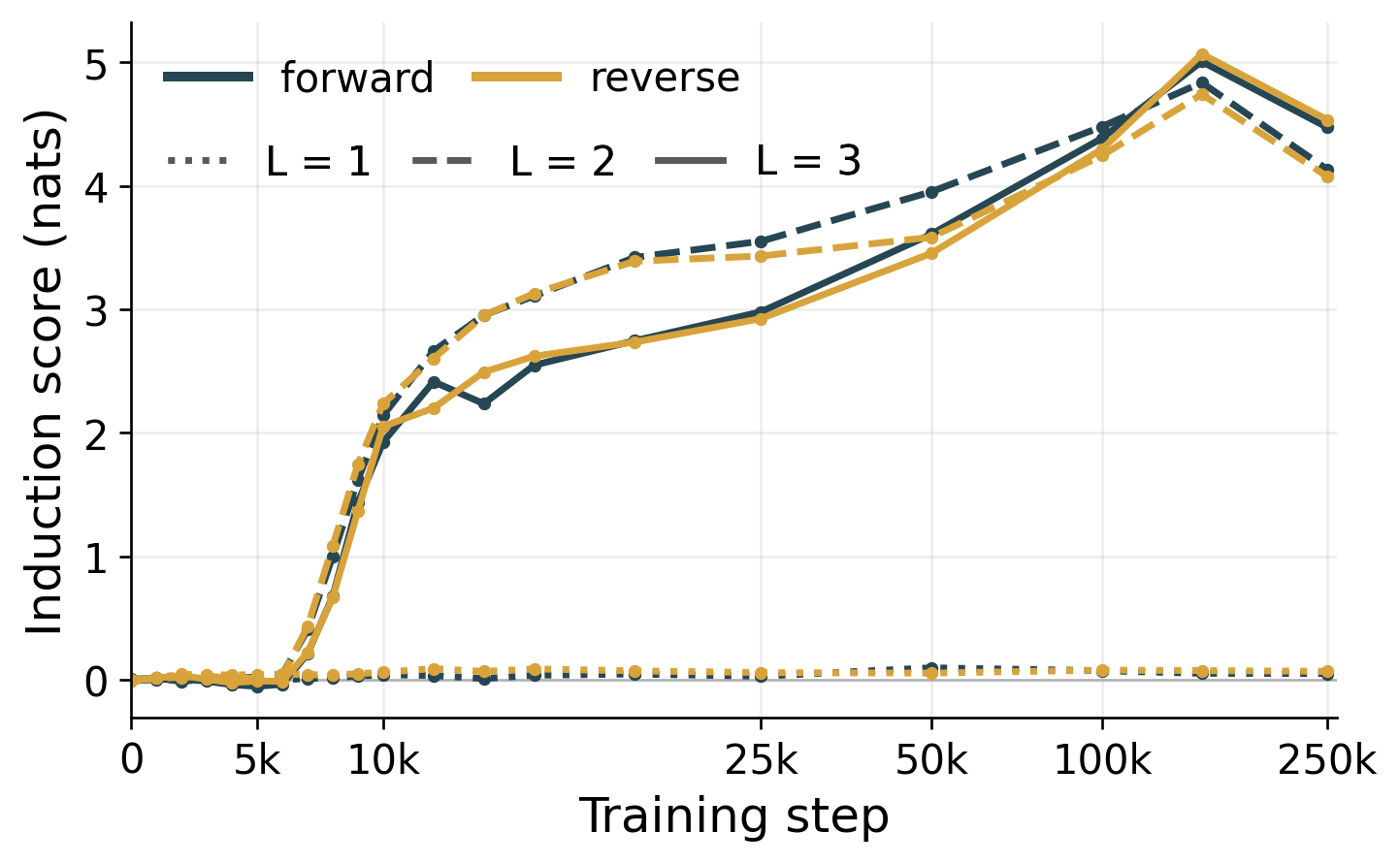}
\caption{DLM induction is direction-symmetric.}
\label{fig:direction-symmetry}
\end{subfigure}
\caption{Induction phase change. The induction score is the log-probability gain on repeated versus control sequences, in nats. Scores are three-seed averages and the step axis is symlog.}
\end{figure}

\subsection{DLM Induction Is Bidirectional}
Induction appears abruptly during training and requires at least two layers, as shown in Figure~\ref{fig:phase-change}. AR and DLM models with one layer stay near zero induction score, while L2 and L3 models transition from near-zero to strong induction after roughly 8-24k training steps, with the DLMs transitioning earlier than the matched AR models. This result is consistent with the expected structure of an induction circuit: one layer is enough to write local neighbour features, but a later layer is needed to use those features to retrieve the matching token~\citep{elhage2021mathematical,olsson2022context}.

Once induction emerges, it is nearly direction-symmetric in the DLM. As shown in Figure~\ref{fig:direction-symmetry}, DLMs perform approximately the same on forward and reverse induction, with nearly identical scores at the end of training (e.g., $4.13 \pm 0.17$ forward vs.\ $4.07 \pm 0.25$ reverse nats at L2). This shows the models retrieve the answer equally well from either direction. An AR model has no corresponding reverse setting because of its causal attention mask.

\begin{figure}[t]
\centering
\includegraphics[width=\linewidth]{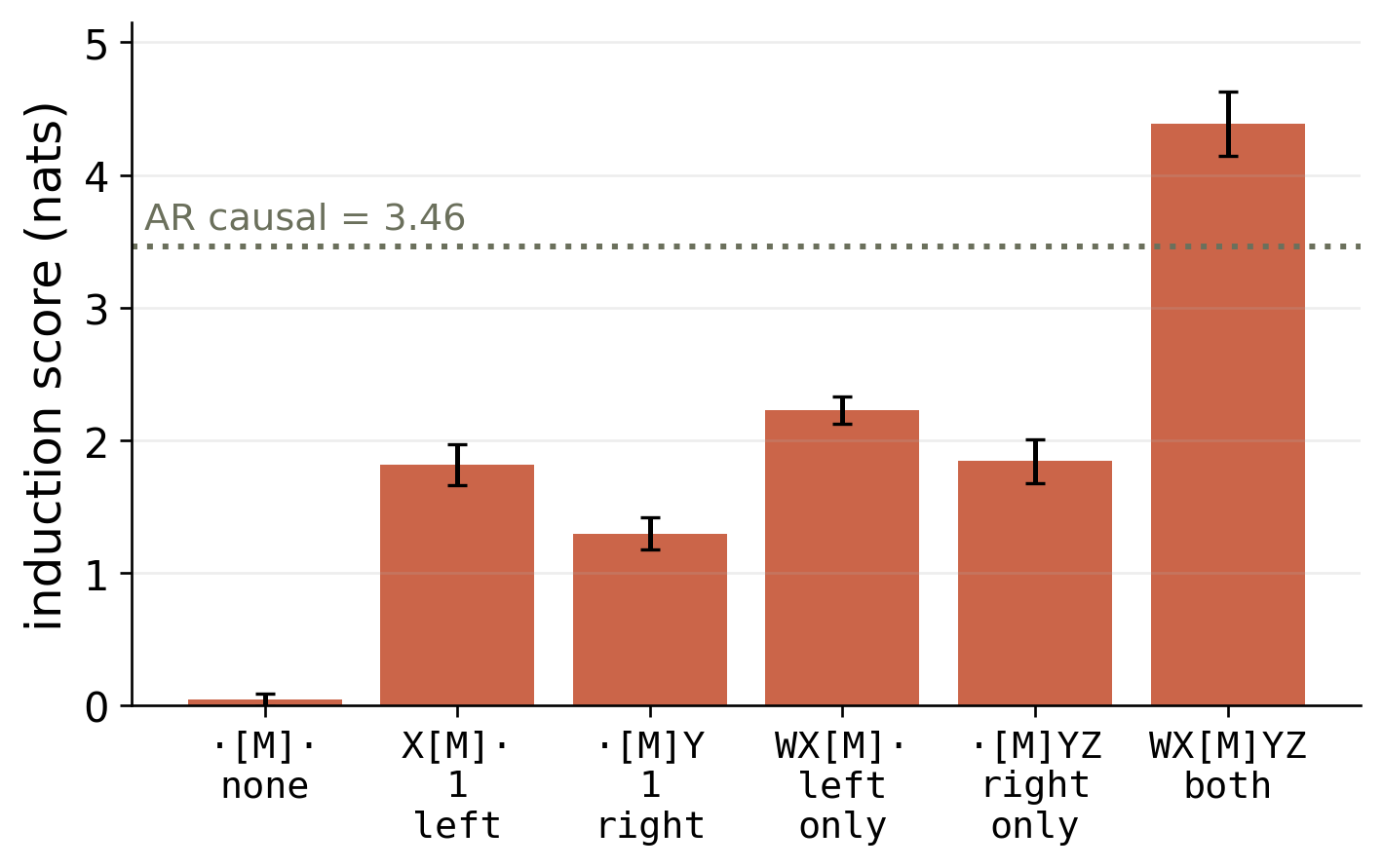}
\caption{Context access at L2, averaged over three seeds. The DLM's advantage over AR appears when both sides of the masked token are visible, and the same pattern holds at L3. Axis schematics are abbreviated from the true $\pm 4$ window.}
\label{fig:cue-capability}
\end{figure}

The DLM has stronger induction when it can see both sides of the masked token (Figure~\ref{fig:cue-capability}). To test this, we control which of the mask's neighbours stay visible within four tokens on each side: none, a single left or right neighbour, the four left, the four right, or all eight. Everything outside this window stays visible, including the whole source copy, and extra masks at distant positions keep the total number of masks equal across conditions, so only the local context around the mask varies.

In the left-only setting, which matches the information available to an AR model, the DLM scores at or below the AR baseline ($2.23$ vs.\ $3.46$ nats at L2). The DLM has stronger induction only when both sides are visible, with a score close to the sum of the two one-sided scores ($4.39$ at L2). This shows that its advantage comes from seeing both sides of the mask rather than from a stronger one-sided induction mechanism.

\subsection{The Circuit Splits Into Previous-Token and Next-Token Pathways}
In order to localize the first part of the circuit, we label layer-0 heads by whether they attend to the previous token or the next token around the mask, and then test their causal role with mean ablation. Previous-token heads attend from the mask to the token on its left, while next-token heads attend from the mask to the token on its right.

\begin{figure}[t]
\centering
\begin{subfigure}{\linewidth}
\centering
\includegraphics[width=\linewidth]{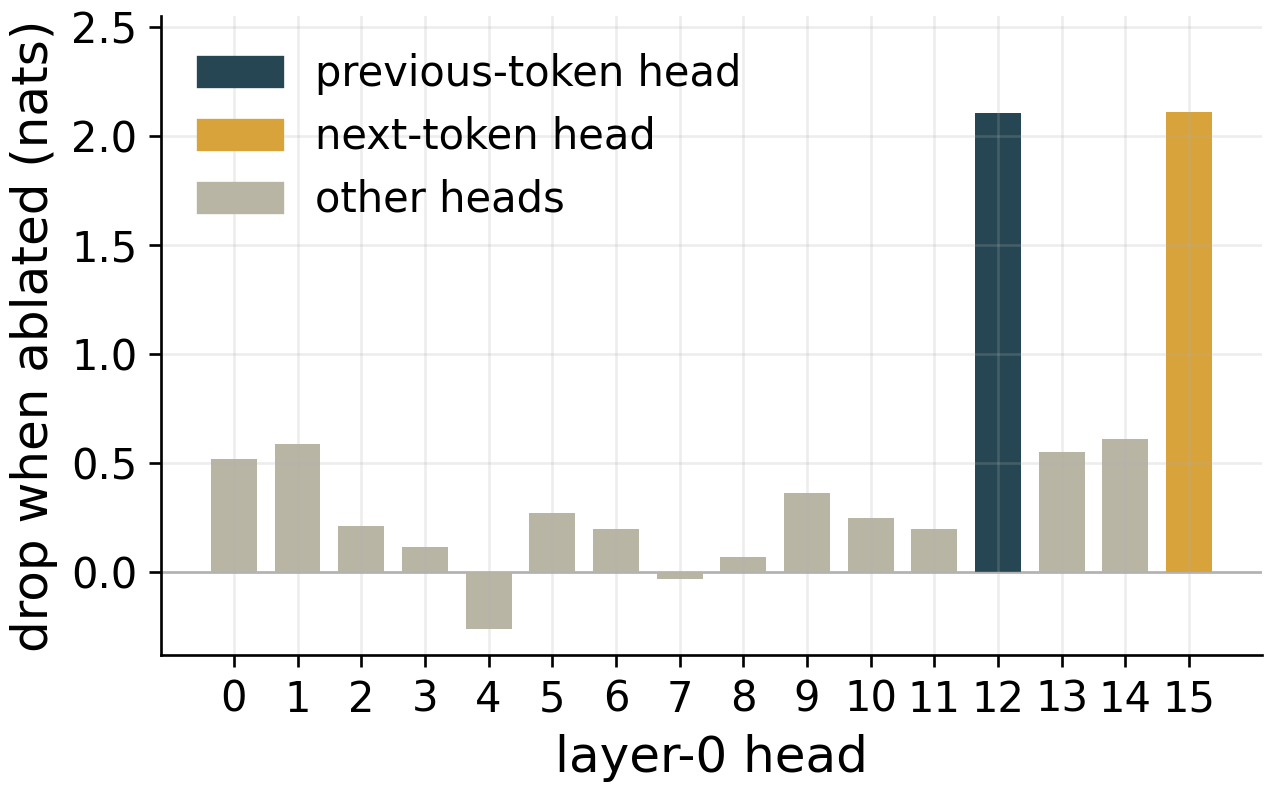}
\caption{The previous-token and next-token heads dominate the layer-0 mean-ablation profile.}
\label{fig:layer0-ablation}
\end{subfigure}
\par\smallskip
\begin{subfigure}{\linewidth}
\centering
\includegraphics[width=\linewidth]{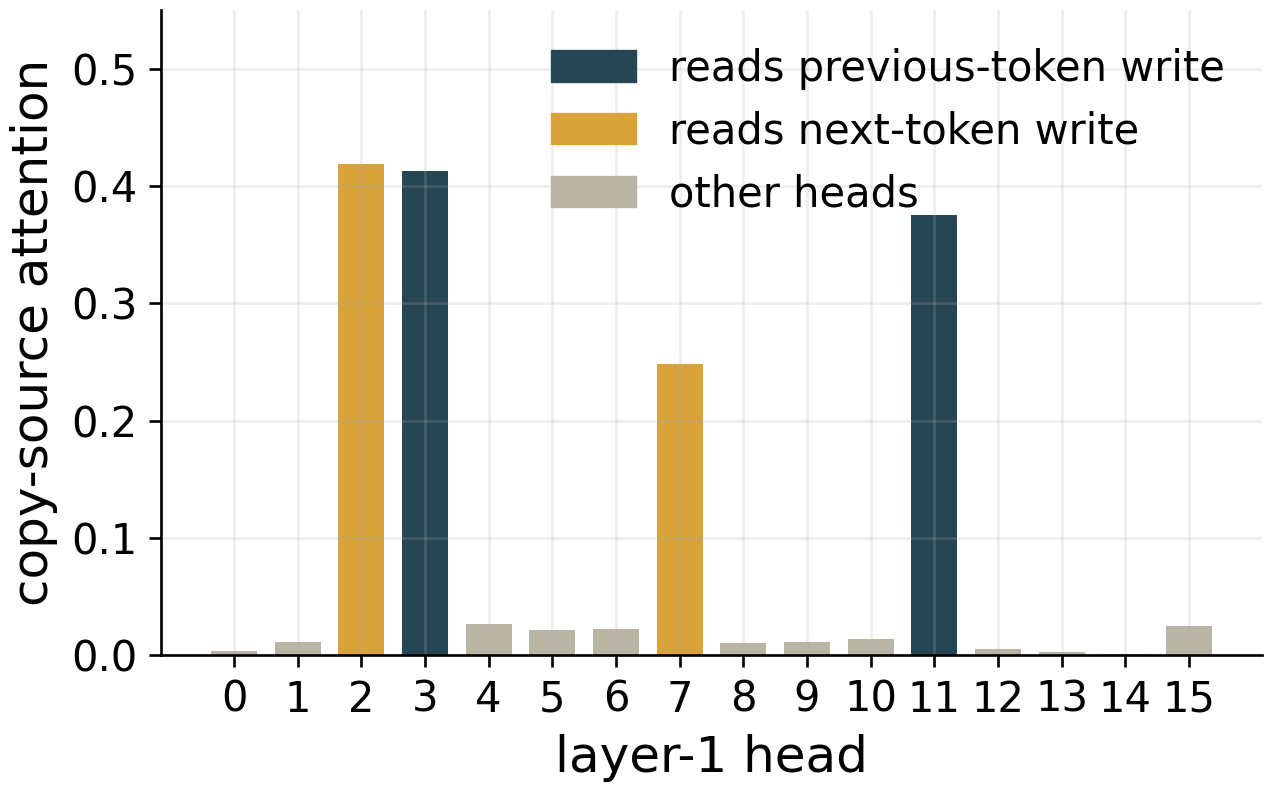}
\caption{A group of layer-1 induction heads attends from the mask to the source answer position.}
\label{fig:layer1-copy-source}
\end{subfigure}
\caption{Circuit localization in one representative DLM run. Across the other DLM runs, head indices differ but the same pattern appears.}
\end{figure}

Ablating each layer-0 head individually singles out one dominant head of each type. In the representative run shown in Figure~\ref{fig:layer0-ablation}, removing the previous-token head or the next-token head causes the largest drops in induction score, while removing any other head produces much smaller changes. This indicates that bidirectional induction starts by writing two local facts into the residual stream: what is immediately to the left of the masked token and what is immediately to its right.

The second step of the circuit appears in the next layer. As shown in Figure~\ref{fig:layer1-copy-source}, a few layer-1 heads attend strongly from the mask to the source answer position in the other copy. We identify these as induction heads: they use the neighbour information written at layer 0 to find the source answer and copy it back to the masked position. We then label each induction head by which layer-0 pathway it reads, determined with QK decomposition.

% Table 1: six-model replication. Requires \usepackage{booktabs,makecell,graphicx}.
\begin{table*}[t]
\centering
\small
\resizebox{\width}{!}{%
\begin{tabular}{llcccccc}
\toprule
depth & run & \makecell{forward\\score\\(nats)} & \makecell{reverse\\score\\(nats)} & \makecell{previous-token\\drop\\(nats)} & \makecell{next-token\\drop\\(nats)} & \makecell{\# induction\\heads} & \makecell{source\\attention\\range} \\
\midrule
L=2 & A & 3.59 & 3.70 & 2.10 & 2.11 & 4 & 0.25-0.42 \\
L=2 & B & 3.54 & 3.82 & 2.07 & 1.62 & 4 & 0.25-0.43 \\
L=2 & C & 3.88 & 3.84 & 2.47 & 1.88 & 4 & 0.34-0.56 \\
\addlinespace
L=3 & A & 4.73 & 4.85 & 3.02 & 3.30 & 5 & 0.44-0.63 \\
L=3 & B & 4.67 & 4.67 & 3.30 & 2.57 & 4 & 0.56-0.70 \\
L=3 & C & 4.51 & 4.48 & 3.00 & 2.16 & 4 & 0.53-0.65 \\
\bottomrule
\end{tabular}%
}
\caption{Replication of the two-pathway circuit across the six folded no-LayerNorm DLMs used for mechanistic analysis. Drops come from ablating one head at a time, and source attention is measured from the mask to the source answer position for the selected induction heads.}
\label{tab:replication}
\end{table*}

The same circuit replicates across depths and training seeds. Table~\ref{tab:replication} lists the corresponding heads for the six folded DLMs (three L2 and three L3 models). Every run has one dominant previous-token head, one dominant next-token head, and a group of layer-1 induction heads whose source attention stands clearly apart from the rest ($0.25$-$0.70$ vs.\ $\approx 0.03$ for the next-highest head). Note that the two per-head ablation drops can sum to more than the full induction score because both ablations disrupt the same downstream copy step. The specific head indices change across runs, but the organization of the circuit is stable. At L3, the final layer adds a weaker group of induction-like heads that read either the layer-1 induction heads or the layer-0 neighbour heads, but the layer-0 to layer-1 circuit remains the dominant path.

We next test whether the two pathways operate independently, using conflict prompts. For the query window \(\mathtt{X\,[M]\,Y}\), we provide two source windows: \(\mathtt{X\,A\,Q}\) matches only on the left and supports answer \(\mathtt{A}\), while \(\mathtt{J\,B\,Y}\) matches only on the right and supports answer \(\mathtt{B}\). The model assigns the two answers nearly equal probability ($\log p(\mathtt{A}) - \log p(\mathtt{B})$ of $-0.01$ to $+0.32$ across the six DLMs, vs.\ $+2.1$ to $+2.8$ when both windows support \(\mathtt{A}\)). Ablating one of the two layer-0 heads tips the prediction toward the other pathway: removing the previous-token head shifts it toward \(\mathtt{B}\), and removing the next-token head shifts it toward \(\mathtt{A}\). This shows that either pathway alone can steer the output.

The two-pathway description is a simplification of a slightly wider local circuit. Besides the dominant heads for $i-1$ and $i+1$, we also find weaker layer-0 heads that attend to $i-2$ and $i+2$. To measure how much local context is used, we construct sequences where only a short segment around the masked position matches the source, instead of the whole sequence, and vary the segment length. The score rises from $0.03$ nats with no matching neighbours to $2.28$ with one matching neighbour on each side and $4.29$ with two, while longer segments add a negligible amount. This suggests an effective induction window of five tokens, although models with more heads and layers may learn a wider one. Ablating a $\pm 2$ head reduces the score by only $0.4$-$0.7$ nats, compared with about $2$ nats for a $\pm 1$ head. Including the $\pm 2$ matches nearly doubles the score because the extra context sharpens the induction heads' attention to the source.

\begin{figure}[hbt]
\centering
\includegraphics[width=\linewidth]{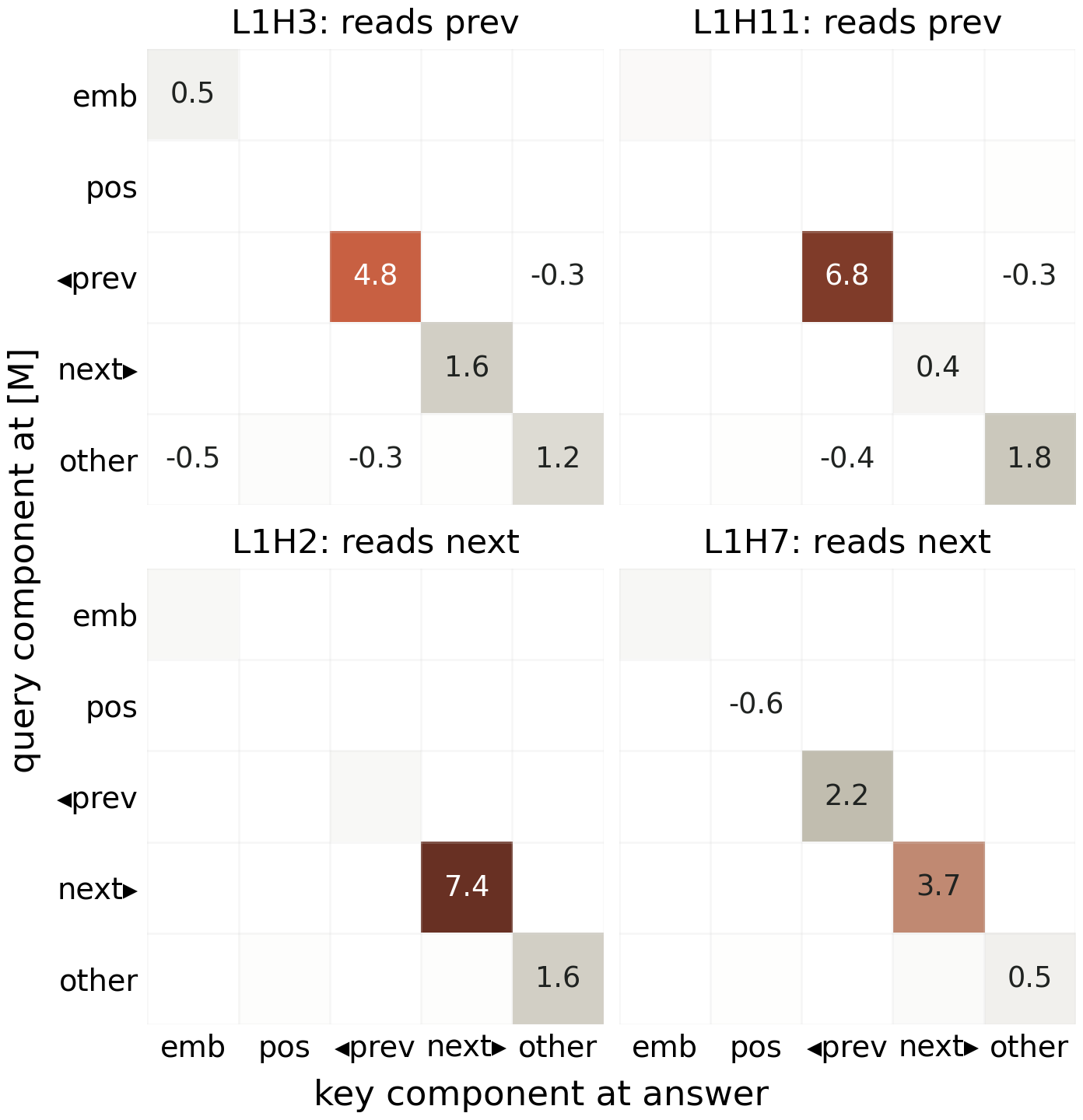}
\caption{QK mechanics in representative induction heads. Each matrix decomposes the attention logit from the mask token \(\mathtt{[M]}\) to the source answer into query-component by key-component terms. The largest term is the one where the head's own layer-0 pathway supplies both the query and the key.}
\label{fig:qk}
\end{figure}

\subsection{Weight Analysis Reveals the QK and OV Mechanics of the Circuit}

We then examine whether the localized heads implement the expected computation in their weights. For each induction head, we decompose the attention logit from the masked position to the source answer into QK contributions from residual-stream components (token embedding, positional embedding, layer-0 head outputs). This lets us identify which earlier component supplies the query at the mask and which component supplies the key at the source position.

The QK decomposition shows that induction heads find the source by matching the same local-neighbour feature in the query and the key. As shown in Figure~\ref{fig:qk}, previous-token induction heads get their largest QK contribution from the previous-token layer-0 head at both the masked position and the source position. Next-token induction heads show the same pattern for the next-token layer-0 head. Across all selected induction heads in the six DLMs, the term where the head's own layer-0 pathway supplies both the query and the key reaches $2.7$-$8.3$ attention logits, while every other term stays below $2.4$.

\begin{figure}[hbt]
\centering
\includegraphics[width=\linewidth]{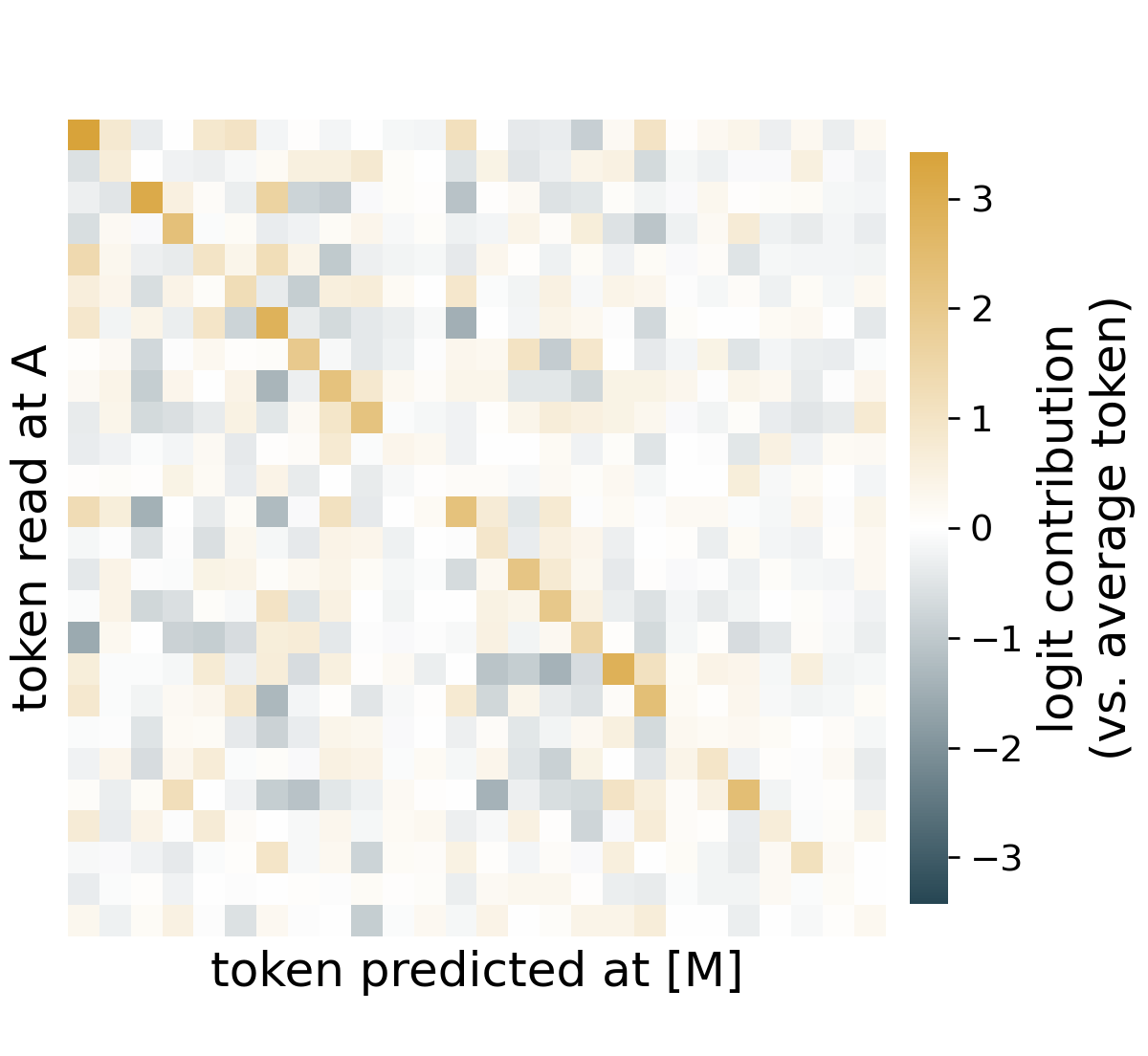}
\caption{OV mechanics in one representative induction head, evaluated on a fixed random sample of 26 vocabulary tokens. Attending to token $t$ at the source tends to raise the logit of $t$ at the mask. Logits are relative to the average vocabulary token.}
\label{fig:ov}
\end{figure}

This shows that the model uses both Q-composition and K-composition: the layer-0 neighbour head helps build the query at the mask token \(\mathtt{[M]}\) and the key at the answer token. This differs from the standard AR induction-head mechanism, where the current token is visible and can build the query from its embedding, so the composition is mainly into the key at the source position~\citep{elhage2021mathematical,olsson2022context}. In a DLM, every masked position carries the same embedding of the mask token \(\mathtt{[M]}\), so the query has to come from what layer-0 heads write at the mask.

The OV path explains what happens once a head attends to the source position. As shown in Figure~\ref{fig:ov}, the effective token-to-token map of an induction head is diagonal on a fixed random sample of 26 vocabulary tokens: attending to token $t$ at the source tends to raise the logit of token $t$ at the mask. Over the full vocabulary, this induction head makes the copied token the top-1 logit for about one in five source tokens.

The weight analysis supports the same mechanism identified by ablation. Layer-0 heads write local neighbour features, QK uses those features to select the matching source position and OV increases the logit of the attended token at the masked position.

\subsection{DLMs Learn an Implicit Timestep}
\begin{figure}[t]
\centering
\includegraphics[width=\linewidth]{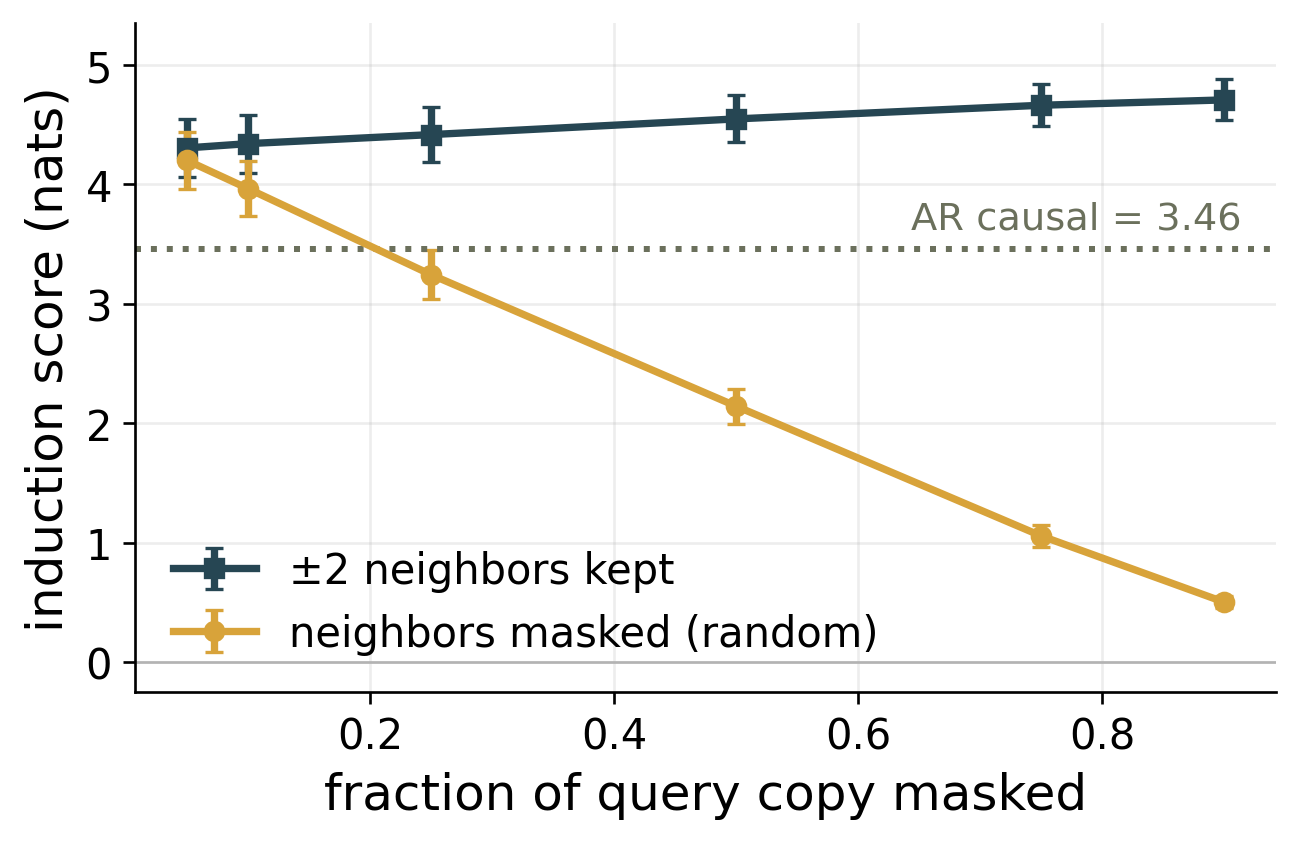}
\caption{Induction under corruption at L2, averaged over three seeds. Induction remains strong under high global mask rates when the local window is visible. The same pattern holds at L3.}
\label{fig:corruption}
\end{figure}

We find that the induction circuit relies on nearby tokens, not on the overall corruption level. The DLM keeps a high induction score under heavy masking as long as the tokens near the mask stay visible (Figure~\ref{fig:corruption}): with the $\pm 2$ neighbours protected, the score even rises slightly as masking reaches $90\%$ (from $4.31$ to $4.71$ at L2). When the nearby tokens are masked instead, the score collapses to about $0.5$. This means the models are robust to the corruption level as long as the local window contains the relevant information.

Although induction is local, the DLM still internally represents the global mask rate. To test this, we train a linear probe to predict the fraction of masked tokens from the residual stream at a single position. At the input embeddings, the probe cannot predict it ($R^2 \approx 0$), as a single token carries no information about the global mask rate. After layer-0 attention, the same probe reaches high $R^2$ in all six DLMs ($0.82$-$0.91$), at both masked and visible positions. This shows that the model computes a global corruption-level feature even though it is not given an explicit timestep embedding~\citep{ou2025your,zheng2025masked}. Since the mask fraction increases with the timestep under the cosine schedule, this feature serves as an implicit timestep.

\begin{figure}[t]
\centering
\begin{subfigure}{\linewidth}
\centering
\includegraphics[width=\linewidth]{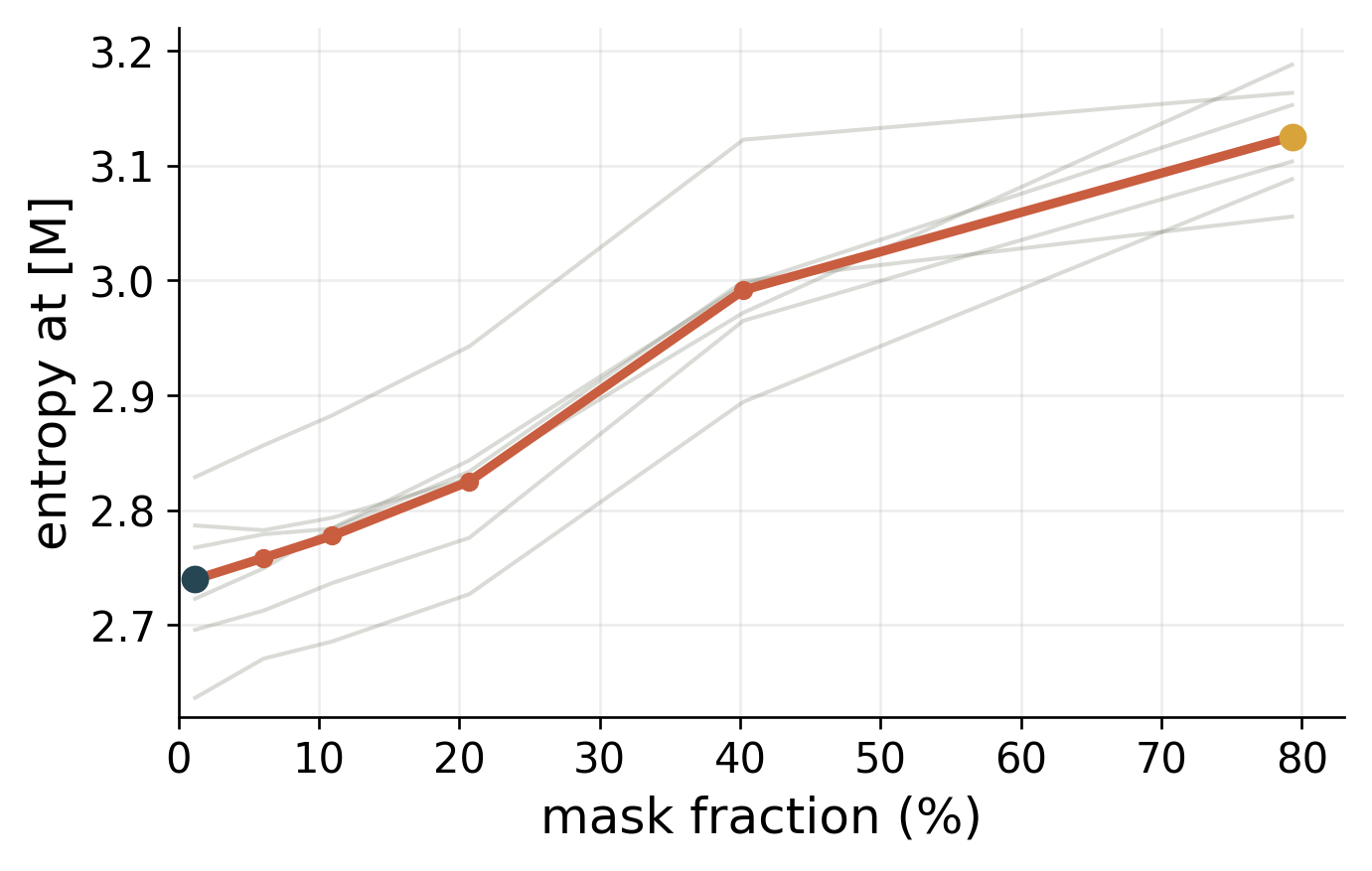}
\caption{With the local context fixed, adding distant masks raises prediction entropy at the masked position.}
\label{fig:timestep-entropy}
\end{subfigure}
\par\smallskip
\begin{subfigure}{\linewidth}
\centering
\includegraphics[width=\linewidth]{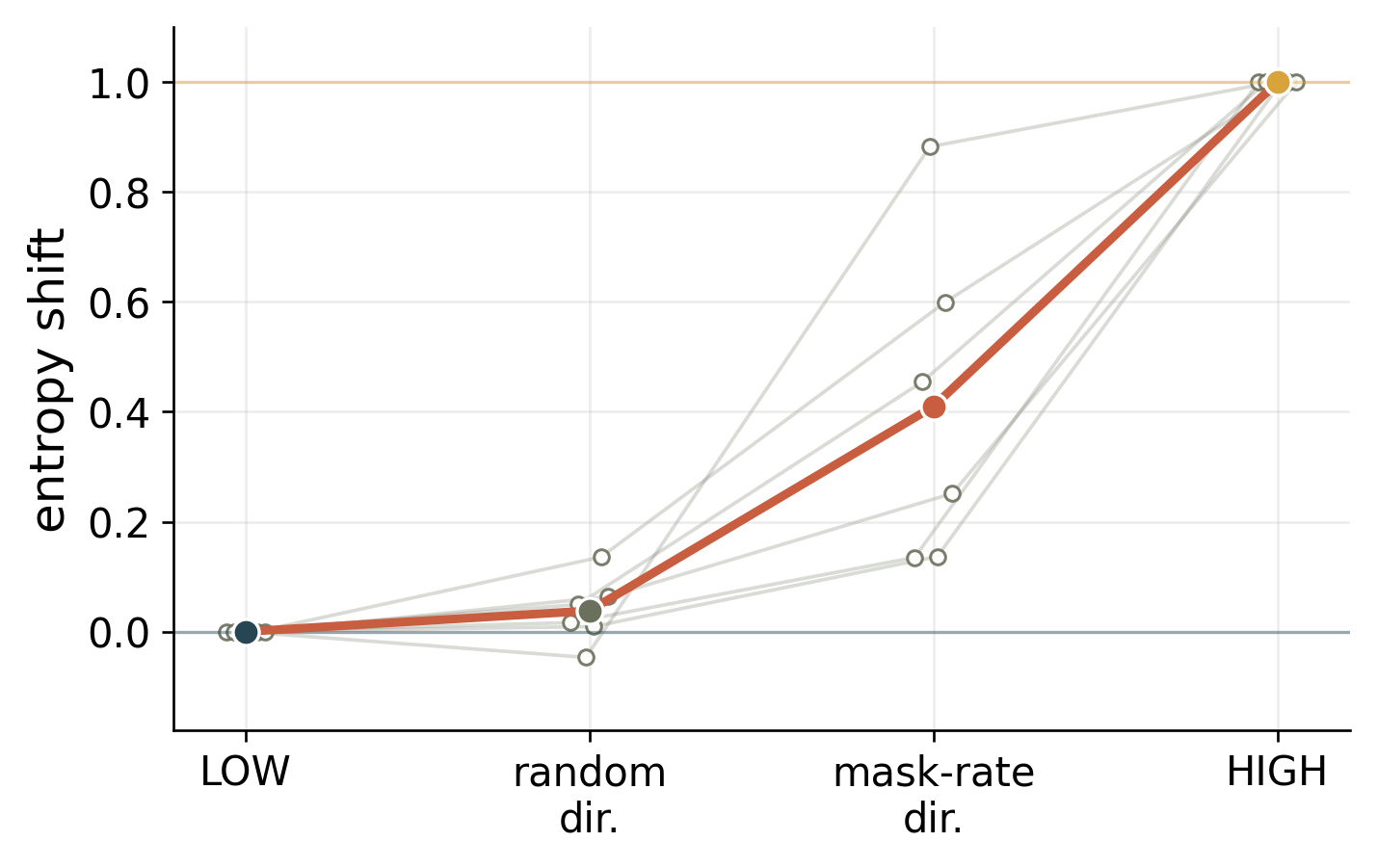}
\caption{Patching the mask-rate direction recovers part of the LOW-to-HIGH entropy shift (normalized so LOW is 0 and HIGH is 1), while an equal-size random-direction patch has little effect.}
\label{fig:timestep-patch}
\end{subfigure}
\caption{The implicit timestep. Gray lines show individual DLMs and the colored line the six-model mean.}
\end{figure}

The mask-rate feature seems to affect predictions. We test this by keeping the visible tokens around a masked position fixed and adding extra masks only at distant positions. As shown in Figure~\ref{fig:timestep-entropy}, the entropy of the prediction at the masked position rises by $0.29$-$0.45$ nats across the six DLMs as the global mask fraction increases. Because the local context is unchanged, this suggests that the model uses the global corruption level when making predictions.

Finally, we test whether this mask-rate representation is causally used. We patch the layer-0 mask-rate direction from a high-corruption run into the same position in a low-corruption run. As shown in Figure~\ref{fig:timestep-patch}, this patch raises entropy toward the high-corruption condition in all six DLMs, recovering $13$-$88\%$ of the low-to-high entropy gap (mean $\approx 40\%$), while an equal-size random-direction patch has little effect. We expect only partial recovery as the patch changes the layer-0 residual only at the predicted position, and later attention layers can recompute the true mask rate from the other positions. Still, the patch shifts entropy in the expected direction in every model, showing that the model not only computes the mask rate but uses it as an implicit timestep.

\section{Limitations}
We analyze small attention-only transformers and their folded no-LayerNorm variants. This simplifies circuit analysis but leaves open if the mechanisms appear in the original LayerNorm models or larger DLMs with MLPs and more layers.

We only study absorbing-mask DLMs, where corrupted positions are replaced by the mask token \(\mathtt{[M]}\). Other diffusion objectives may behave differently. In particular, uniform-noise DLMs do not use a mask token~\citep{austin2021structured,lou2023discrete}, so they may not learn the same local induction circuit or the same implicit timestep.

%Finally, our induction task is synthetic and controlled. Random-token repeated sequences remove semantic and frequency effects, which is useful for isolating the circuit, but this does not show that the same circuit explains all in-context learning behaviour in natural language. Testing this mechanism on richer tasks remains an important next step.

Finally, our induction task is synthetic and controlled. Random-token repeated sequences isolate the induction circuit by eliminating semantic and frequency effects, but this does not show that the same circuit underlies in-context learning in natural language. Testing this mechanism on richer tasks remains an important next step.
\section{Conclusion}
We present a controlled mechanistic study of induction in matched AR and absorbing-mask DLMs. We show that DLMs learn a bidirectional circuit: previous-token and next-token pathways feed later induction heads, whose QK and OV weights retrieve and copy the source answer token.

The DLM advantage over AR comes from access to both sides of the masked token rather than a stronger one-sided mechanism. We also find that DLMs compute and causally use the global mask rate as an implicit timestep, even without an explicit timestep embedding.

Our work is a step toward understanding the mechanisms behind in-context learning beyond AR models, an important setting as other classes of language models are increasingly adopted.

\clearpage
\bibliography{references}

@article{olsson2022context,
  title={In-context learning and induction heads},
  author={Olsson, Catherine and Elhage, Nelson and Nanda, Neel and Joseph, Nicholas and DasSarma, Nova and Henighan, Tom and Mann, Ben and Askell, Amanda and Bai, Yuntao and Chen, Anna and others},
  journal={arXiv preprint arXiv:2209.11895},
  year={2022}
}

@article{elhage2021mathematical,
  title={A mathematical framework for transformer circuits},
  author={Elhage, Nelson and Nanda, Neel and Olsson, Catherine and Henighan, Tom and Joseph, Nicholas and Mann, Ben and Askell, Amanda and Bai, Yuntao and Chen, Anna and Conerly, Tom and others},
  journal={Transformer Circuits Thread},
  volume={1},
  number={1},
  pages={12},
  year={2021}
}

@article{wang2022interpretability,
  title={Interpretability in the wild: a circuit for indirect object identification in gpt-2 small},
  author={Wang, Kevin and Variengien, Alexandre and Conmy, Arthur and Shlegeris, Buck and Steinhardt, Jacob},
  journal={arXiv preprint arXiv:2211.00593},
  year={2022}
}

@article{goldowsky2023localizing,
  title={Localizing model behavior with path patching},
  author={Goldowsky-Dill, Nicholas and MacLeod, Chris and Sato, Lucas and Arora, Aryaman},
  journal={arXiv preprint arXiv:2304.05969},
  year={2023}
}

@article{austin2021structured,
  title={Structured denoising diffusion models in discrete state-spaces},
  author={Austin, Jacob and Johnson, Daniel D and Ho, Jonathan and Tarlow, Daniel and Van Den Berg, Rianne},
  journal={Advances in neural information processing systems},
  volume={34},
  pages={17981--17993},
  year={2021}
}

@article{gulrajani2023likelihood,
  title={Likelihood-based diffusion language models},
  author={Gulrajani, Ishaan and Hashimoto, Tatsunori B},
  journal={Advances in Neural Information Processing Systems},
  volume={36},
  pages={16693--16715},
  year={2023}
}

@article{lou2023discrete,
  title={Discrete diffusion modeling by estimating the ratios of the data distribution},
  author={Lou, Aaron and Meng, Chenlin and Ermon, Stefano},
  journal={arXiv preprint arXiv:2310.16834},
  year={2023}
}

@article{nie2026large,
  title={Large language diffusion models},
  author={Nie, Shen and Zhu, Fengqi and You, Zebin and Zhang, Xiaolu and Ou, Jingyang and Hu, Jun and Zhou, Jun and Lin, Yankai and Wen, Ji-Rong and Li, Chongxuan},
  journal={Advances in Neural Information Processing Systems},
  volume={38},
  pages={50608--50646},
  year={2026}
}

@article{khanna2025mercury,
  title={Mercury: Ultra-fast language models based on diffusion},
  author={Khanna, Samar and Kharbanda, Siddhant and Li, Shufan and Varma, Harshit and Wang, Eric and Birnbaum, Sawyer and Luo, Ziyang and Miraoui, Yanis and Palrecha, Akash and Ermon, Stefano and others},
  journal={arXiv e-prints},
  pages={arXiv--2506},
  year={2025}
}

@article{wang2026dlm,
  title={DLM-Scope: Mechanistic Interpretability of Diffusion Language Models via Sparse Autoencoders},
  author={Wang, Xu and Jiang, Bingqing and Wan, Yu and Yang, Baosong and Kong, Lingpeng and Zou, Difan},
  journal={arXiv preprint arXiv:2602.05859},
  year={2026}
}

@article{dai2026revealing,
  title={Revealing the Attention Floating Mechanism in Masked Diffusion Models},
  author={Dai, Xin and Huang, Pengcheng and Liu, Zhenghao and Wang, Shuo and Yan, Yukun and Xiao, Chaojun and Gu, Yu and Yu, Ge and Sun, Maosong},
  journal={arXiv preprint arXiv:2601.07894},
  year={2026}
}

@article{kong2026mechanism,
  title={Mechanism Shift During Post-training from Autoregressive to Masked Diffusion Language Models},
  author={Kong, Injin and Lee, Hyoungjoon and Jo, Yohan},
  journal={arXiv preprint arXiv:2601.14758},
  year={2026}
}

@article{garg2025masked,
  title={Masked Diffusion Models are Secretly Learned-Order Autoregressive Models},
  author={Garg, Prateek and Kohli, Bhavya and Sarawagi, Sunita},
  journal={arXiv preprint arXiv:2511.19152},
  year={2025}
}

@article{sahoo2024simple,
  title={Simple and effective masked diffusion language models},
  author={Sahoo, Subham S and Arriola, Marianne and Schiff, Yair and Gokaslan, Aaron and Marroquin, Edgar and Chiu, Justin T and Rush, Alexander and Kuleshov, Volodymyr},
  journal={Advances in Neural Information Processing Systems},
  volume={37},
  pages={130136--130184},
  year={2024}
}

@article{shi2024simplified,
  title={Simplified and generalized masked diffusion for discrete data},
  author={Shi, Jiaxin and Han, Kehang and Wang, Zhe and Doucet, Arnaud and Titsias, Michalis},
  journal={Advances in neural information processing systems},
  volume={37},
  pages={103131--103167},
  year={2024}
}

@inproceedings{ou2025your,
  title={Your absorbing discrete diffusion secretly models the conditional distributions of clean data},
  author={Ou, Jingyang and Nie, Shen and Xue, Kaiwen and Zhu, Fengqi and Sun, Jiacheng and Li, Zhenguo and Li, Chongxuan},
  booktitle={International Conference on Learning Representations},
  volume={2025},
  pages={64972--65009},
  year={2025}
}

@inproceedings{zheng2025masked,
  title={Masked diffusion models are secretly time-agnostic masked models and exploit inaccurate categorical sampling},
  author={Zheng, Kaiwen and Chen, Yongxin and Mao, Hanzi and Liu, Ming-Yu and Zhu, Jun and Zhang, Qinsheng},
  booktitle={International Conference on Learning Representations},
  volume={2025},
  pages={63186--63227},
  year={2025}
}

@inproceedings{nie2025scaling,
  title={Scaling up masked diffusion models on text},
  author={Nie, Shen and Zhu, Fengqi and Du, Chao and Pang, Tianyu and Liu, Qian and Zeng, Guangtao and Lin, Min and Li, Chongxuan},
  booktitle={International Conference on Learning Representations},
  volume={2025},
  pages={82974--82997},
  year={2025}
}

@inproceedings{gong2025scaling,
  title={Scaling diffusion language models via adaptation from autoregressive models},
  author={Gong, Shansan and Agarwal, Shivam and Zhang, Yizhe and Ye, Jiacheng and Zheng, Lin and Li, Mukai and An, Chenxin and Zhao, Peilin and Bi, Wei and Han, Jiawei and others},
  booktitle={International Conference on Learning Representations},
  volume={2025},
  pages={5046--5073},
  year={2025}
}

@article{ye2025dream,
  title={Dream 7b: Diffusion large language models},
  author={Ye, Jiacheng and Xie, Zhihui and Zheng, Lin and Gao, Jiahui and Wu, Zirui and Jiang, Xin and Li, Zhenguo and Kong, Lingpeng},
  journal={arXiv preprint arXiv:2508.15487},
  year={2025}
}

@inproceedings{arriola2025block,
  title={Block diffusion: Interpolating between autoregressive and diffusion language models},
  author={Arriola, Marianne and Gokaslan, Aaron and Chiu, Justin and Yang, Zhihan and Qi, Zhixuan and Han, Jiaqi and Sahoo, Subham and Kuleshov, Volodymyr},
  booktitle={International Conference on Learning Representations},
  volume={2025},
  pages={50726--50753},
  year={2025}
}

@article{kim2025train,
  title={Train for the worst, plan for the best: Understanding token ordering in masked diffusions},
  author={Kim, Jaeyeon and Shah, Kulin and Kontonis, Vasilis and Kakade, Sham and Chen, Sitan},
  journal={arXiv preprint arXiv:2502.06768},
  year={2025}
}

@article{elhage2022toy,
  title={Toy models of superposition},
  author={Elhage, Nelson and Hume, Tristan and Olsson, Catherine and Schiefer, Nicholas and Henighan, Tom and Kravec, Shauna and Hatfield-Dodds, Zac and Lasenby, Robert and Drain, Dawn and Chen, Carol and others},
  journal={arXiv preprint arXiv:2209.10652},
  year={2022}
}

@article{templeton2026scaling,
  title={Scaling monosemanticity: Extracting interpretable features from claude 3 sonnet},
  author={Templeton, Adly and Conerly, Tom and Marcus, Jonathan and Lindsey, Jack and Bricken, Trenton and Chen, Brian and Pearce, Adam and Citro, Craig and Ameisen, Emmanuel and Jones, Andy and others},
  journal={arXiv preprint arXiv:2605.29358},
  year={2026}
}

@inproceedings{huben2024sparse,
  title={Sparse autoencoders find highly interpretable features in language models},
  author={Huben, Robert and Cunningham, Hoagy and Smith, Logan and Ewart, Aidan and Sharkey, Lee},
  booktitle={International Conference on Learning Representations},
  volume={2024},
  pages={7827--7845},
  year={2024}
}

@article{conmy2023towards,
  title={Towards automated circuit discovery for mechanistic interpretability},
  author={Conmy, Arthur and Mavor-Parker, Augustine and Lynch, Aengus and Heimersheim, Stefan and Garriga-Alonso, Adri{\`a}},
  journal={Advances in Neural Information Processing Systems},
  volume={36},
  pages={16318--16352},
  year={2023}
}

@article{nanda2023progress,
  title={Progress measures for grokking via mechanistic interpretability},
  author={Nanda, Neel and Chan, Lawrence and Lieberum, Tom and Smith, Jess and Steinhardt, Jacob},
  journal={arXiv preprint arXiv:2301.05217},
  year={2023}
}

@misc{odonoghue2026diffusiongemma,
    title = {DiffusionGemma: 4x faster text generation},
    author = {O'Donoghue, Brendan and Flennerhag, Sebastian},
    year = {2026},
    month = jun,
    howpublished = {\url{https://blog.google/innovation-and-ai/
    technology/developers-tools/diffusion-gemma-faster-text-
    generation/}},
    note = {Google Blog}
  }

@article{heimersheim2024you,
  title={You can remove GPT2's LayerNorm by fine-tuning},
  author={Heimersheim, Stefan},
  journal={arXiv preprint arXiv:2409.13710},
  year={2024}
}

@misc{nanda2022transformerlens,
    title = {TransformerLens},
    author = {Neel Nanda and Joseph Bloom},
    year = {2022},
    howpublished = {\url{https://github.com/TransformerLensOrg/TransformerLens}},
}

@article{penedo2024fineweb,
  title={The fineweb datasets: Decanting the web for the finest text data at scale},
  author={Penedo, Guilherme and Kydl{\'\i}{\v{c}}ek, Hynek and Lozhkov, Anton and Mitchell, Margaret and Raffel, Colin and Von Werra, Leandro and Wolf, Thomas and others},
  journal={Advances in Neural Information Processing Systems},
  volume={37},
  pages={30811--30849},
  year={2024}
}

@article{radford2019language,
  title={Language models are unsupervised multitask learners},
  author={Radford, Alec and Wu, Jeffrey and Child, Rewon and Luan, David and Amodei, Dario and Sutskever, Ilya and others},
  journal={OpenAI blog},
  volume={1},
  number={8},
  pages={9},
  year={2019}
}

@article{loshchilov2017decoupled,
  title={Decoupled weight decay regularization},
  author={Loshchilov, Ilya and Hutter, Frank},
  journal={arXiv preprint arXiv:1711.05101},
  year={2017}
}

@article{noy2023experimental,
  title={Experimental evidence on the productivity effects of generative artificial intelligence},
  author={Noy, Shakked and Zhang, Whitney},
  journal={Science},
  volume={381},
  number={6654},
  pages={187--192},
  year={2023},
  publisher={American Association for the Advancement of Science}
}

@article{peng2023impact,
  title={The impact of ai on developer productivity: Evidence from github copilot},
  author={Peng, Sida and Kalliamvakou, Eirini and Cihon, Peter and Demirer, Mert},
  journal={arXiv preprint arXiv:2302.06590},
  year={2023}
}

@article{boiko2023autonomous,
  title={Autonomous chemical research with large language models},
  author={Boiko, Daniil A and MacKnight, Robert and Kline, Ben and Gomes, Gabe},
  journal={Nature},
  volume={624},
  number={7992},
  pages={570--578},
  year={2023},
  publisher={Nature Publishing Group UK London}
}

@article{bereska2024mechanistic,
  title={Mechanistic interpretability for AI safety--a review},
  author={Bereska, Leonard and Gavves, Efstratios},
  journal={arXiv preprint arXiv:2404.14082},
  year={2024}
}

@inproceedings{lee2022coauthor,
  title={Coauthor: Designing a human-ai collaborative writing dataset for exploring language model capabilities},
  author={Lee, Mina and Liang, Percy and Yang, Qian},
  booktitle={Proceedings of the 2022 CHI conference on human factors in computing systems},
  pages={1--19},
  year={2022}
}

@article{belinkov2022probing,
  title={Probing classifiers: Promises, shortcomings, and advances},
  author={Belinkov, Yonatan},
  journal={Computational Linguistics},
  volume={48},
  number={1},
  pages={207--219},
  year={2022}
}

@inproceedings{artetxe2022role,
  title={On the role of bidirectionality in language model pre-training},
  author={Artetxe, Mikel and Du, Jingfei and Goyal, Naman and Zettlemoyer, Luke and Stoyanov, Veselin},
  booktitle={Findings of the Association for Computational Linguistics: EMNLP 2022},
  pages={3973--3985},
  year={2022}
}

@article{patel2022bidirectional,
  title={Bidirectional language models are also few-shot learners},
  author={Patel, Ajay and Li, Bryan and Rasooli, Mohammad Sadegh and Constant, Noah and Raffel, Colin and Callison-Burch, Chris},
  journal={arXiv preprint arXiv:2209.14500},
  year={2022}
}

@article{mueller2026quest,
  title={The quest for the right mediator: Surveying mechanistic interpretability for nlp through the lens of causal mediation analysis},
  author={Mueller, Aaron and Brinkmann, Jannik and Li, Millicent and Marks, Samuel and Pal, Koyena and Prakash, Nikhil and Rager, Can and Sankaranarayanan, Aruna and Sharma, Arnab Sen and Sun, Jiuding and others},
  journal={Computational Linguistics},
  volume={52},
  number={1},
  pages={331--378},
  year={2026},
  publisher={MIT Press 255 Main Street, 9th Floor, Cambridge, Massachusetts 02142, USA~…}
}

\end{document}